\documentclass{mva_style}
\usepackage{graphicx}
\usepackage{color}

\usepackage{amsmath,amsfonts,amssymb}
\usepackage{graphicx}
\usepackage[colorlinks=true, allcolors=blue]{hyperref}

\usepackage{multirow}

\finalcopy 

\begin{document}
\title{Any-scale Object Detection using Arbitrary-scaled Images}

\author{
  Kazutoshi Akita\\
  Toyota Technological Institute\\
  {\tt sd21501@toyota-ti.ac.jp}\\
  \and
  Norimichi Ukita\\
  Toyota Technological Institute\\
  {\tt ukita@toyota-ti.ac.jp}\\
}

\maketitle


\section*{\centering Abstract}
\textit{
This paper proposes any-scale object detection using arbitrary-scale super-resolution for continuously rescaling object images, while general multi-scale object detection uses discretely rescaled appearance representations.
However, a na\"{i}ve usage of super-resolution produces many false-positive detections if many super-resolution images are independently fed into an object detector.
Our method suppresses these false positives by predicting scale proposal maps, each of which represents a set of pixels appropriate for each super-resolution scale.
}


\section{Introduction}
\label{section: introduction}

Object detection is one of the most fundamental tasks required by various applications.
Many issues make object detection difficult, e.g., occlusion, cluttered background, and motion blur.
Among all of these, this paper focuses on the difficulty of detecting objects observed with different scales, which is called multi-scale object detection.
Most multi-scale object detection is based on pyramid representations (e.g., SSD~\cite{SSD}, FPN~\cite{DBLP:conf/cvpr/LinDGHHB17}, EfficientDet~\cite{DBLP:conf/cvpr/TanPL20}, SNIPER~\cite{SNIPER}).

While the pyramid representation is essentially effective both for upsampling~\cite{SSD,DBLP:conf/cvpr/LinDGHHB17,DBLP:conf/cvpr/TanPL20} and downsampling~\cite{SNIPER}, image and feature scales are discretized, such as $\times \frac{1}{4}$, $\times \frac{1}{2}$, $\times 1$, $\times 2$, and $\times 4$.
Either of these discretized scales approximates each object scale in images.
For example, large, medium, and small objects are scaled to $\times \frac{1}{4}$, $\times 1$, and, $\times 4$, respectively, for matching the scales of these objects.
While such approximate representations may fail to detect intermediate-scale objects, e.g., a medium-large object should be scaled to $\times 1.5$, it is impractical to prepare all fine-grained discretized scale representations due to the memory issue.

\begin{figure}[t]
\centering
  \includegraphics[width=\linewidth]{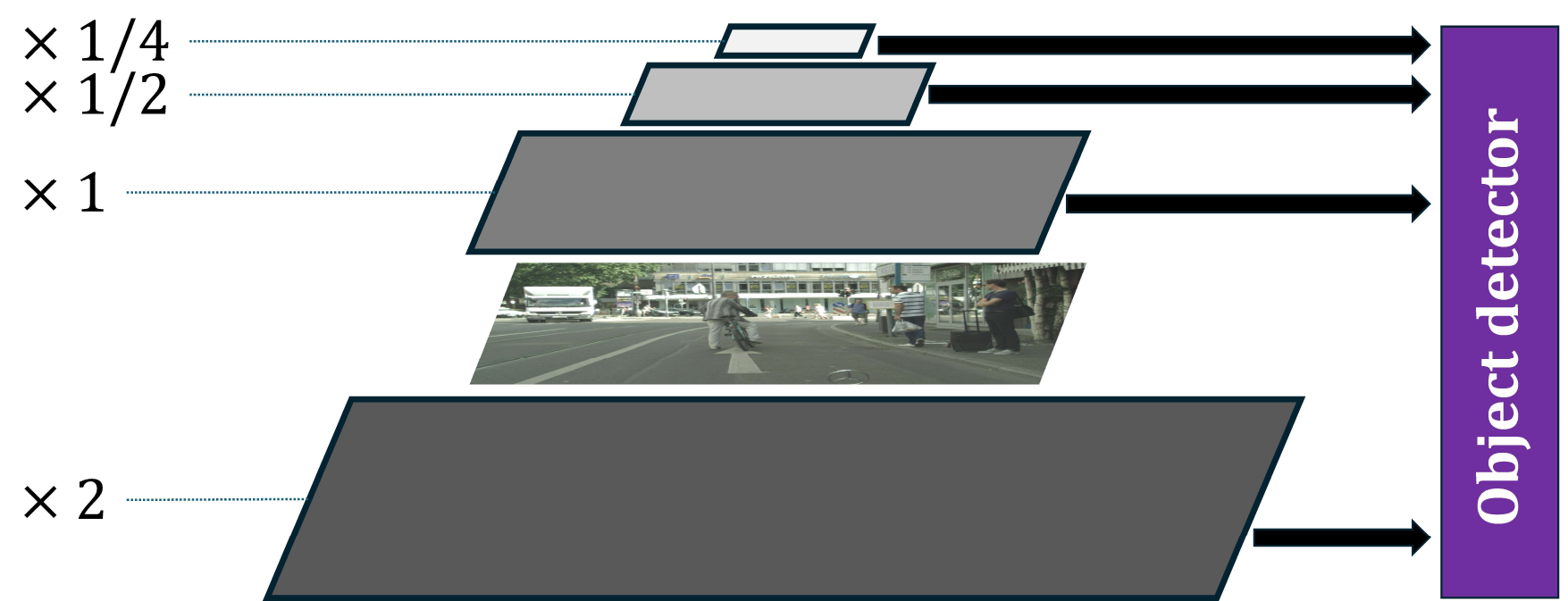}
  (a) {\bf Multi-scale} detection using discretized pyramid\\
  \vspace*{2mm}
  \includegraphics[width=\linewidth]{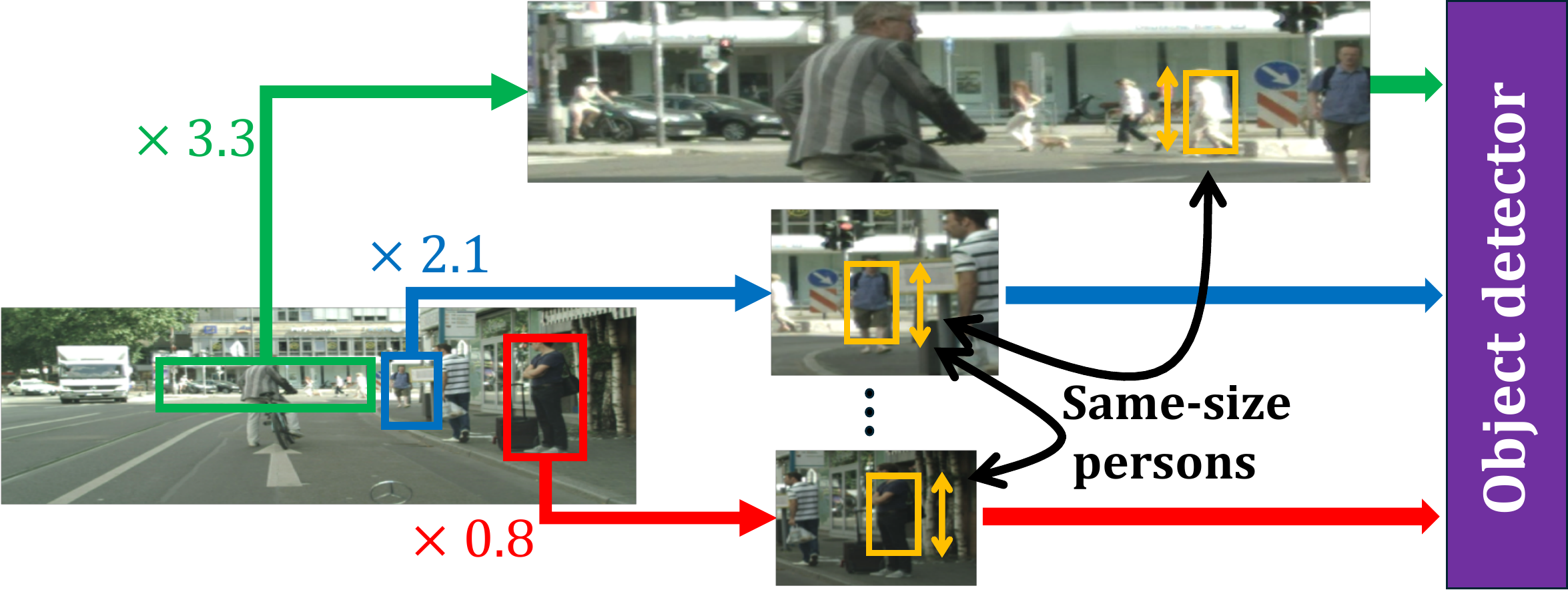}
  (b) {\bf Any-scale} object detection\\ using arbitrary-scale super-resolution\\
  \vspace*{2mm}
  \caption{Our proposed any-scale object detection shown in (b). Gray boxes in (a) indicate downsampled and upsampled images/features.}
  \label{fig: teaser}
  \vspace*{-4mm}
\end{figure}

To cope with the discretized scales, this paper proposes to employ arbitrary-scale image scaling (i.e., Arbitrary-Scale Super-Resolution, ASSR) for any-scale object detection.
While ASSR supports an object detector to find all scale objects as many as possible to increase true-positive detections, object detection in images with various scales raises not only true positives but also false positives.
This is because, in a particular scale, ASSR unsuccessfully produces local appearances that are accidentally similar to object appearances in other scales.

To suppress these false positives, this paper proposes arbitrary-scale region proposals.
Unlike a set of general region proposals~\cite{DBLP:conf/iccv/Girshick15} including all scales, our arbitrary-scale region proposals are separated into those in each scaled image reconstructed by ASSR.
In Fig.~\ref{fig: teaser} (b), for example, green, blue, and red boxes are region proposals for super-resolving by factors $\times 3.1$, $\times 2.2$, and $\times 1.6$, respectively.
While the state-of-the-art method~\cite{DBLP:journals/access/AkitaU23} closest to our method provides region proposals in each scaled image, the scales are discretized.

This paper proposes any-scale object detection using ASSR with region proposals optimized for any-scale images.
The contributions of this paper are as follows:
\begin{itemize}
\item This paper extends multi-scale object detection to any-scale object detection in which ASSR is optimized for the object detection task. 
\item Our region proposal network connected to the object detection network is trained jointly for all scales.
This joint training is essential to efficiently generalize the network even with training data (i.e., annotated region proposals) of sparsely-and-discontinuously distributed scales.
\end{itemize}


\section{Related Work}
\label{section: related}

\subsection{Super-resolution}

In many super-resolution methods~\cite{DBLP:conf/cvpr/LedigTHCCAATTWS17,DBLP:conf/cvpr/ZhangTKZ018,DBPN,RBPN,haris2020space,timofte2018ntire,gu2019aim,nah2019ntire,zhang2020ntire}, each SR model is limited to be SR with a single integer scaling factor, such as $\times 2$, $\times 3$, and $\times 4$.
On the other hand, ASSR~\cite{metasr,Wang21,Vas23,Wang23arb,Son21,Wang19fusion} allows for any scaling factors via a single SR model.

In several ASSR methods~\cite{metasr,Wang21,Vas23,Wang23arb}, appropriate upsampling filters are predicted.
ASSR can also be achieved by blending SR images of integer scale factors~\cite{Son21,Wang19fusion}. 

Another approach for ASSR is based on Local Implicit Image Function (LIIF~\cite{liif}). 
LIIF constructs an implicit function that estimates the RGB value of each pixel from its pixel position and its nearest latent code obtained from an input LR image. LTE~\cite{lte} and Thera~\cite{becker2025thera} predict Fourier components by the LIIF-based framework. CiaoSR~\cite{ciaosr} and CLIT~\cite{clit} aggregate multiple latent codes beyond the nearest one. UltraSR~\cite{ultrasr} and IPE~\cite{ipe} replace the input pixel position of LIIF with the embedded one. 

In addition to the ability for arbitrary scaling, recurrentLTE~\cite{recurrentLTE} allows for a tradeoff between the SR quality and the computational cost.
This tradeoff allows us to control the runtime of computationally expensive ASSR methods depending on the application.
For example, object detection for intelligent vehicles needs to be fast to predict risky moving trajectories.

\subsection{Multi-scale Object Detection}

Pyramid representations~\cite{SSD,DBLP:conf/cvpr/LinDGHHB17,DBLP:conf/cvpr/TanPL20,SNIP,SNIPER} allow us to detect multi-scale objects.
Many feature pyramid-based methods are proposed (e.g., SSD~\cite{SSD}, FPN~\cite{DBLP:conf/cvpr/LinDGHHB17}, BiFPN~\cite{DBLP:conf/cvpr/TanPL20}, PAFPN~\cite{DBLP:conf/cvpr/LiuQQSJ18}, NAS-FPN~\cite{DBLP:conf/cvpr/GhiasiLL19}, AugFPN~\cite{DBLP:conf/cvpr/GuoFZXP20}, and SuperYOLO~\cite{DBLP:journals/tgrs/ZhangLXFLD23}), while image pyramids are also employed for multi-scale object detection (e.g., SNIP~\cite{SNIP}, SNIPER~\cite{SNIPER}, and PIIP~\cite{DBLP:conf/nips/Zhu0W0DGL0D24}).
Since attention mechanisms in Transformer are useful for many tasks, including object detection (e.g., DETR~\cite{DBLP:conf/eccv/CarionMSUKZ20}, Dynamic Head~\cite{DBLP:conf/cvpr/DaiCX0LY021}, and ViT-CoMer~\cite{DBLP:conf/cvpr/XiaWLHS24}), they can be integrated with the pyramid representations for further improvement (e.g., \cite{DBLP:conf/eccv/ZhangZTWHS20,DBLP:conf/cvpr/ChenC0CTKM22,DBLP:journals/nn/MinLL22,DBLP:journals/tgrs/LiuTCWSC24}).

While most of the aforementioned pyramid representations downscale input images and features, it is known that upsampling images and features allows us to improve the detection performance of small objects~\cite{SNIP,SNIPER,DBLP:conf/mva/FujiiAU21,DBLP:conf/mva/KondoUYHSHHHXWLHKLKHKISLLY23}.
However, the methods above upsample images and features degraded by simple interpolation-based upsampling.
To improve the upsampling quality, super resolution (SR) can be used, as proposed in~\cite{DBLP:conf/iccv/NohBLSK19,DBLP:conf/cvpr/ShermeyerE19,DBLP:journals/remotesensing/CourtraiPL20,DBLP:conf/ipas2/AkitaHU20,DBLP:conf/iconip/HarisSU21,DBLP:journals/access/AkitaU23}.
Our proposed method further improves these SR-based methods by ASSR,


\section{Method}
\label{section: method}

\begin{figure}[t]
\centering
  \includegraphics[width=\linewidth]{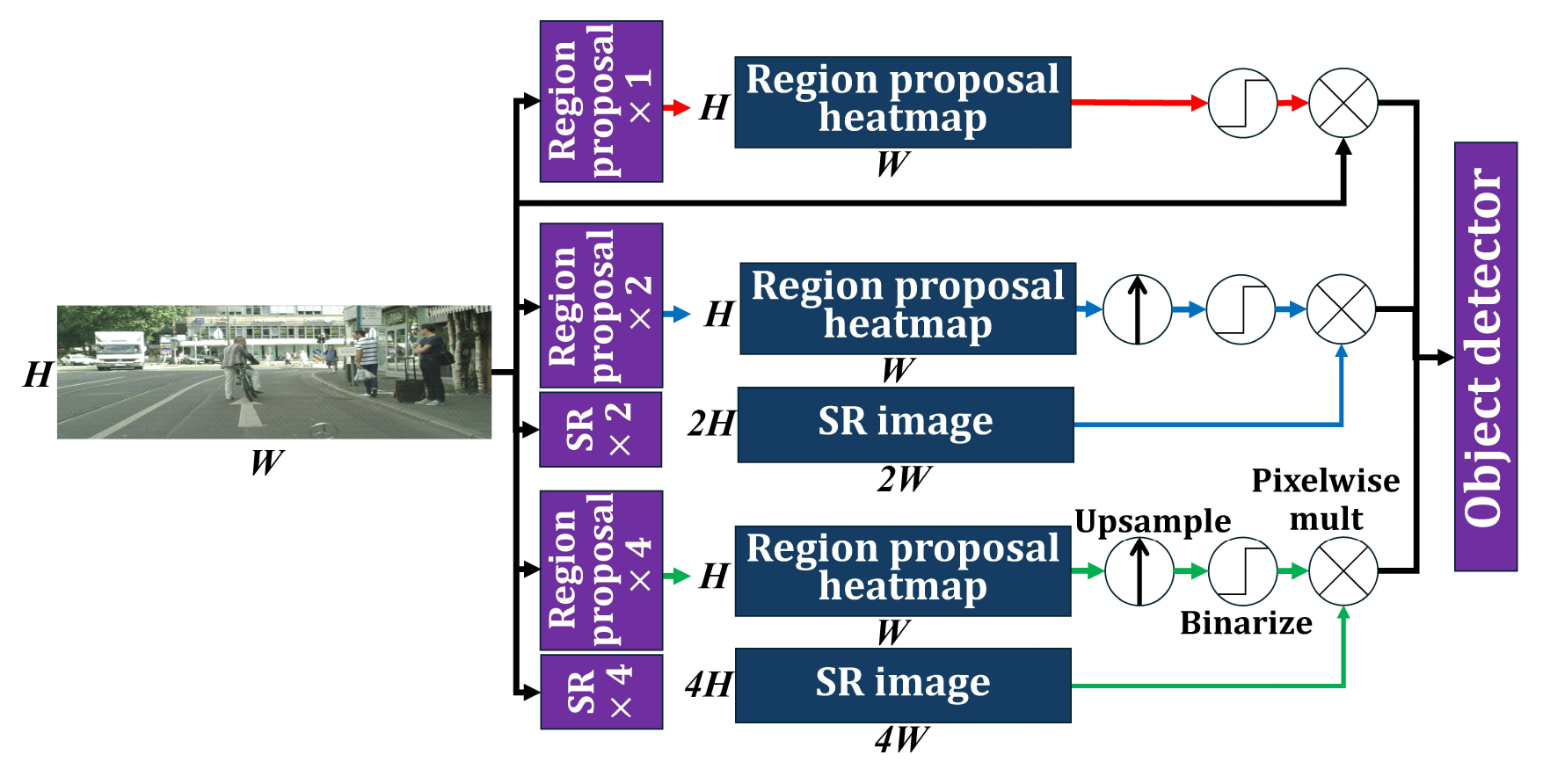}
  \vspace*{-6mm}\\
  (a) With scale-discretized region proposals~\cite{DBLP:journals/access/AkitaU23}\\
  \vspace*{2mm}
  \includegraphics[width=\linewidth]{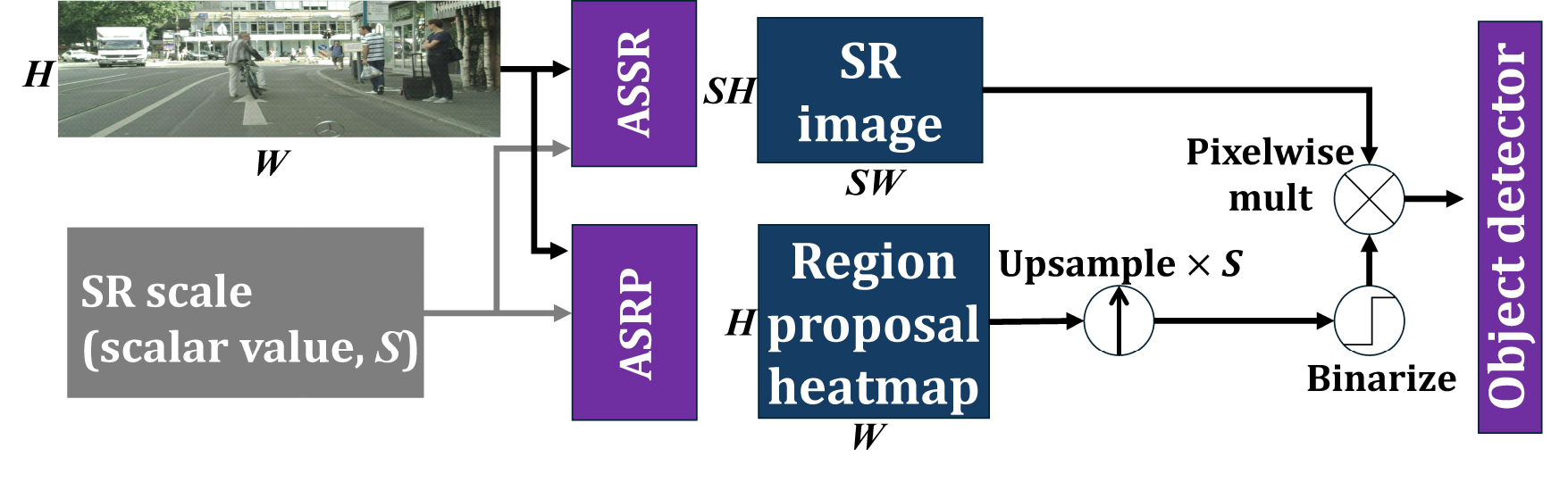}
  \vspace*{-6mm}\\
  (b) With arbitrary-scale region proposals (ours)\\
  \vspace*{2mm}
  \caption{Our any-scale object detection (b), consisting of a single pair of the arbitrary-scale region proposal (ASRP) and ASSR networks, is executed multiple times with any $S \in \mathbb R$. But, multiple region proposals and SR networks are used independently in (a). ASRP is detailed in Fig.~\ref{fig: asrp}.}
  \label{fig: rp}
  \vspace*{-4mm}
\end{figure}

\begin{figure}[t]
\centering
  \includegraphics[width=\linewidth]{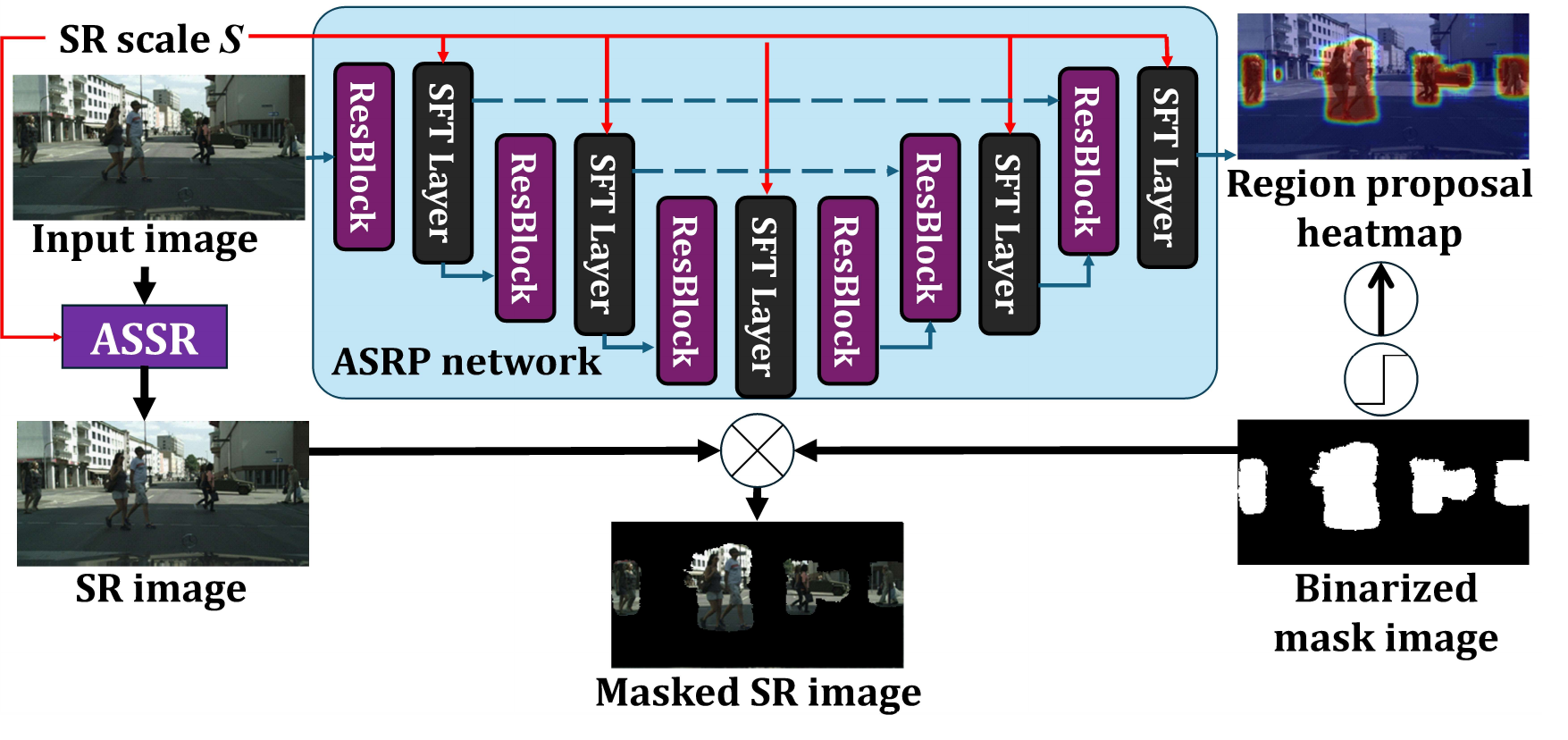}
  \caption{Detail of the proposed ASRP network.}
  \vspace*{-2mm}
  \label{fig: asrp}
  \vspace*{-4mm}
\end{figure}

While the previous method~\cite{DBLP:journals/access/AkitaU23} closest to ours uses region proposals, this method~\cite{DBLP:journals/access/AkitaU23} must have {\bf multiple pairs of SR and region proposal networks} (Fig.~\ref{fig: rp} (a)).
Furthermore, due to the limitation of general SR networks, this method~\cite{DBLP:journals/access/AkitaU23} is limited to {\bf integer scale factors}.
In contrast, our method has {\bf only a pair of ASSR and arbitrary-scale region proposal (ASRP) networks} (Fig.~\ref{fig: rp} (b)).
Furthermore, ASSR allows for {\bf any scale factor}, $S \in \mathbb R$. This flexibility enables finer object-scale alignment, improving detection performance.
In practice, while ASSR allows any continuous scaling factor, our experiments evaluate detection using representative scales in \{0.5, 1.0, 1.5, 2.0, 2.5\}. This choice ensures coverage of the object scale variation while keeping computational cost tractable. Extending to denser scales remains straightforward within our framework.

As shown in Fig.~\ref{fig: rp} (b), an input image and an SR scale $S$ are given to the ASSR network and the ASRP network independently.
Given the input image with $W \times H$ pixels, the SR image with $SW \times SH$ pixels and the region proposal heatmap with $W \times H$ pixels are generated.
Each pixel value of this heatmap is between 0 and 1, which means that this pixel must not be and must be super-resolved by $S$, respectively.
The heatmap is resized to $SW \times SH$ pixels and binarized by thresholding.
This binarized heatmap is regarded as a mask image and is multiplied with the SR image.

\subsection{Arbitrary-scale Region Proposal Network}
\label{subsection: ASRP}

The detail of our ASRP network is shown in Fig.~\ref{fig: asrp}.

\noindent
{\bf Inference:}
The ASRP network takes an input image, $I_{i}$, and the SR factor, $S$.
The ASRP network follows the architecture of the previous region proposal approach~\cite{DBLP:journals/access/AkitaU23}, using Residual blocks~\cite{DBLP:conf/cvpr/HeZRS16}. We extend this design by adding SFT layers~\cite{DBLP:conf/cvpr/WangYDL18} to enable conditioning on the scaling factor $S$.
Each SFT layer takes $S$ to condition the ASRP network to predict the region proposal heatmap, $I^{(S)}_{h}$, highlighting pixels that should be super-resolved by factor $S$.
$I^{(S)}_{h}$ is upscaled by factor $S$ and binarized to make the binarized mask image, $I^{(S)}_{m}$.

$I_{i}$ and $S$ are also fed into ASSR to reconstruct the SR image, $I^{S}_{s}$.
$I^{(S)}_{s}$ is masked by $I^{(S)}_{m}$ to make the masked SR image, $I^{(S)}_{ms}$.
Finally, $I^{(S)}_{ms}$ is fed into an object detector.

\noindent
{\bf Training:}
The ASRP network is trained in a supervised manner.
Given bounding box annotations in object detection datasets, our method quantifies the scale of each object bounding box by its height.
Let $H_{t}$ denote the target height to which all objects are rescaled by ASSR. In our experiments, $H_{t} = 128$ is chosen as a hyperparameter based on empirical experiments.
Each training image is processed so that pixel values inside and outside a bounding box are 1 and 0, respectively.
This processed image is regarded as the ground truth of the region proposal heatmap when its original image and the SR scale $S' = \frac{H_{t}}{H_{i}}$, where $H_{i}$ denotes the object height in the input image, are fed into the ASRP network.
However, if this heatmap is directly used as the training image, the ASRP network overfits with only one object appearance in each training image.
To avoid this overfitting, not only one object but also any other objects, each of whose scale is between $\alpha \frac{H_{t}}{S'}$ and $\beta \frac{H_{t}}{S'}$, also highlighted so that their pixels values are 1 in the heatmap.
The hyperparameters $\alpha=0.7$ and $\beta = 1.3$ are empirically determined to provide sufficient tolerance around the target scale.
Furthermore, the pixel values in the heatmap are blurred, following~\cite{DBLP:journals/access/AkitaU23}.

\subsection{Object Detector}
\label{subsection: detector}

Ideally, only the same-scale objects are fed into an object detector in our any-scale object detection method.
However, general object detectors are trained to detect multi-scale objects.
Therefore, while our method uses such an object detector as a pretrained model, this detector is fine-tuned to detect only objects of the same scale with a height of $H_{t}$.
The detail of this finetuning is described in Sec.~\ref{subsection: exp_details}.

After objects are detected from all the masked SR images, all these objects are merged by a general non-maximum suppression algorithm.


\section{Experiments}
\label{section: experiments}

\subsection{Dataset}
\label{subsection: exp_dataset}

We conducted experiments with the CityScapes dataset~\cite{cityscapes}, which is a car-mounted camera dataset. 
Since this dataset was developed to evaluate instance segmentation methods, the annotations are pixelwise instance labels.
From these pixelwise instance labels, bounding boxes for object detection were generated so that a rectangle circumscribing each instance is regarded as its bounding box.

While 30 object classes are defined in the dataset, most of them are background objects such as ``sky'' and ``road.'' In our experiments, only classes included in the human group (i.e., ``person'' and ``rider'') were used for the object detection task.
The dataset is split into training and test images. The number of training and test images containing human bounding boxes is 2,965 and 492, respectively.
In all the training and test images, 19,801 and 3,975 human bounding boxes are included, respectively.

\begin{table}[t]
\centering
\vspace*{3mm}
\caption{Detection performance comparison. 
(a) DETR. (b) DETR is used in~\cite{DBLP:journals/access/AkitaU23}, in which SR and region proposals with integer SR factors. (c) DETR is used with ASSR~\cite{lte}. (d) DETR is used with ASSR and ASRP (ours).}
\label{table: exp_main_ablation}
\scalebox{0.82}{
\begin{tabular}{l|ccccc} \hline
          & AP  & AP$_{50}$ & AP$_{s}$ & AP$_{m}$ & AP$_{l}$ \\ \hline
    (a) DETR~\cite{DBLP:conf/eccv/CarionMSUKZ20}  & 29.2 & 60.2    & 11.6   & 38.0   & 47.4 \\
    (b) (a)+SR+RDSP~\cite{DBLP:journals/access/AkitaU23} & 30.1 & 61.4    & 13.6   & 38.3   & 47.4 \\
    (c) (a)+ASSR~\cite{lte}   & 30.8 & 59.6    & 14.8   & 38.7   & 47.3 \\
    (d) (c)+ASRP & 32.9 & 63.2    & 19.0   & 41.0   & 47.3 \\ \hline
\end{tabular}
}
\vspace*{6mm}
\centering
\caption{Detection performance comparison. In DETR with ASSR and ASRP, an input image is super-resolved with varying the range of SR factors. (d) in Table~\ref{table: exp_main_asrp} is the same with (d) in Table~\ref{table: exp_main_ablation} (ours).}
\label{table: exp_main_asrp}
\scalebox{0.85}{
\begin{tabular}{l|ccccc} \hline
    Range of SR factors & AP  & AP$_{50}$ & AP$_{s}$ & AP$_{m}$ & AP$_{l}$ \\ \hline
    (a) DETR  & 29.2 & 60.2    & 11.6   & 38.0   & 47.4 \\
    (b) b/w 0.5 and 1.5  & 31.4 & 62.3    & 13.0   & 41.6   & 47.4 \\
    (c) b/w 0.5 and 2.0  & 32.7 & 63.3    & 18.2   & 41.4   & 47.3 \\
    (d) b/w 0.5 and 2.5  & 32.9 & 63.2    & 19.0   & 41.0   & 47.3 \\
    (e) b/w 0.5 and 3.0  & 32.9 & 63.2    & 19.0   & 41.0   & 47.3 \\ \hline
\end{tabular}
}
\end{table}

\begin{figure}[t]
    \centering
    \includegraphics[width=\linewidth]{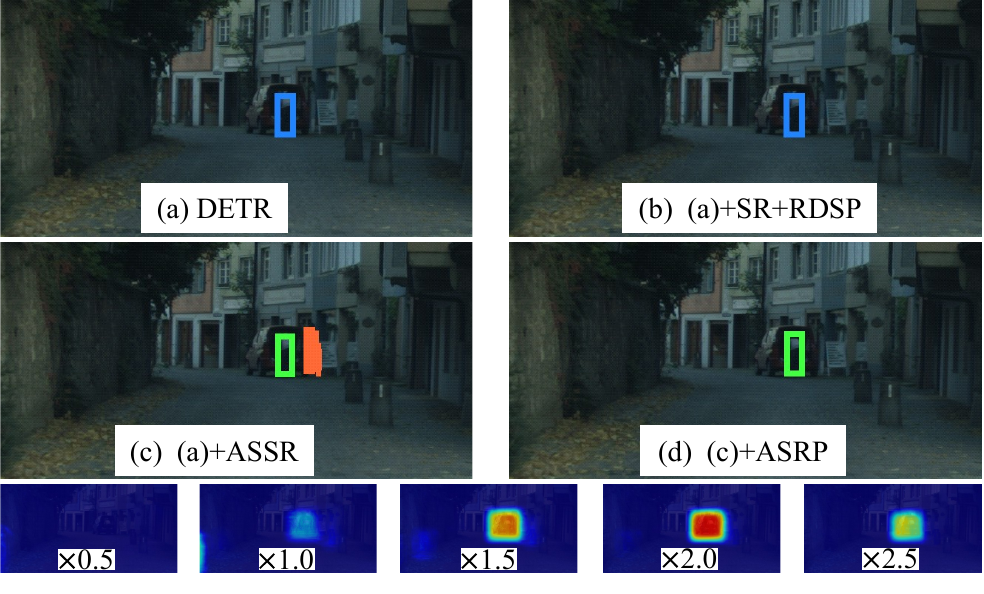} \\
    \includegraphics[width=\linewidth]{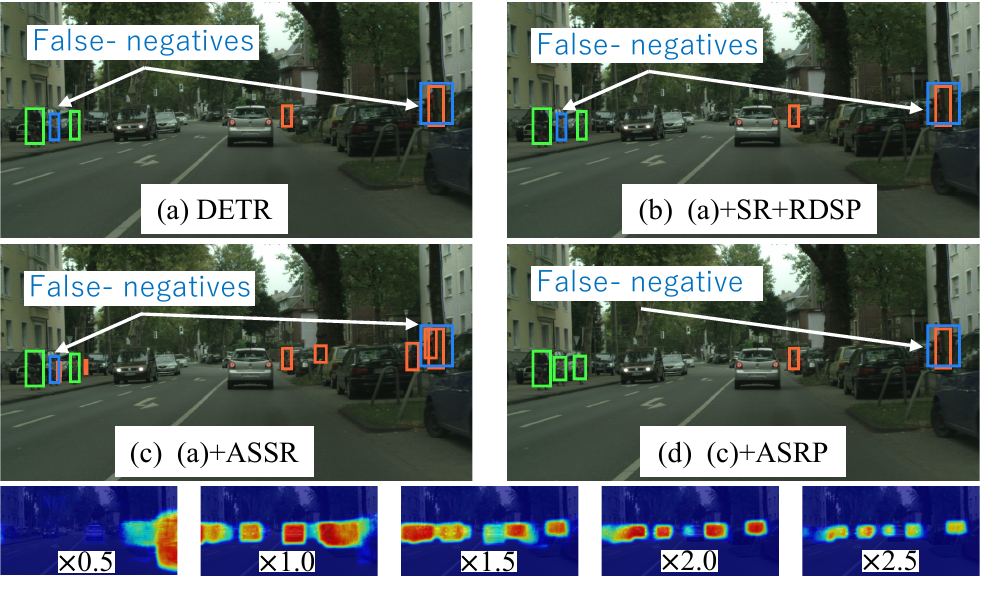}
    \vspace*{-5mm}
    \caption{Visual results. Green, blue, and red boxes indicate true positives, false negatives, and false positives, respectively. Five images at the bottom of each example are the heatmaps of (d).}
    \label{fig: success}
    \vspace*{-5mm}
\end{figure}

\subsection{Implementation Details}
\label{subsection: exp_details}

Since our method can use any object detector, a Transformer-based object detector, DETR~\cite{DBLP:conf/eccv/CarionMSUKZ20}, is used in our experiments as one of the most potent object detectors.
Its official implementation and publicly available model trained with the COCO dataset are used.
For ASSR also, any network can be used.
In our experiments, LTE~\cite{lte} is used.
As with DETR, LTE is implemented with the official code.
Its publicly available model is trained with the DIV2K dataset~\cite{div2k}.

The aforementioned pretrained models of DETR and LTE are concatenated with our ASRP network, which is trained from scratch, as shown in Fig.~\ref{fig: rp} (b).
This joint network is trained with the training set of the CityScapes dataset in an end-to-end manner.
For this joint learning, the whole network is trained to detect only same-scale objects, as described in Sec.~\ref{subsection: detector}. The optimizer used in this learning is RAdam~\cite{DBLP:conf/iclr/LiuJHCLG020} with an initial learning rate of 1e-5 and a weight decay of 1e-5. The learning rate is halved at epoch 300, while the total epochs are 400. The batch size is 8.

In~\cite{DBLP:journals/access/AkitaU23}, in addition to masked SR images, the original input image
is directly fed into the object detector.
All objects detected from this original image and all the masked SR images are merged.
Following~\cite{DBLP:journals/access/AkitaU23}, our method also feeds the original image to the detector.


\subsection{Quantitative Results}
\label{subsection: exp_quantitative}

(a) DETR~\cite{DBLP:conf/eccv/CarionMSUKZ20} is fine-tuned with the CityScapes dataset.
This DETR is trained in its original multi-scale detection manner.
In (b), DETR is used with~\cite{DBLP:journals/access/AkitaU23} using SR and region proposals with integer SR factors
(c) DETR is used with ASSR~\cite{lte} so that DETR takes SR images without region proposals.
(d) 
DETR with ASSR and ASRP
is our method.

As shown in Table~\ref{table: exp_main_ablation}, while the scores of large object detection (i.e., AP$_{l}$) are the almost same in all the methods, our method is superior to the others in small object detection (i.e., AP$_{s}$).


The effect of varying SR factors is shown in Table~\ref{table: exp_main_asrp}.
It is natural that the detection scores improve as the range expands.
This improvement is saturated around in (d) because these ranges sufficiently include all object scales in the test images.

\noindent
{\bf Memory efficiency.}
The number of parameters in (a), (b), (c), and (d) in Table~\ref{table: exp_main_ablation}
are (a) 41.5M, (b) 151.1M, (c) 42.8M, and (d) 78.0M, respectively.
While (d) our method is larger than methods without region proposals (i.e., (a) and (c)), it is smaller than (b) the state-of-the-art method with multiple SR and region proposal networks because our method has only a single pair of ASSR and ASRP networks.


\subsection{Qualitative Results}

Two detection examples of the four methods are shown in Fig.~\ref{fig: success}.
In the upper example, (a) DETR and (b) DETR with SR and RDSP cannot detect any objects.
While (c) DETR with ASSR detects a true positive, a false positive is also detected.
In (d) our method, on the other hand, only a true-positive is detected.
We can also see that the region proposal heatmap is successfully highlighted around the object.

In the lower example, (d) our method detects an object located in the left side in the image, which is a false negative in (a), (b), and (c).
However, two false positives and one false negative cannot be resolved even in (d).
While the heatmaps of different SR scales are different in each example, their overlaps are not small. Improving the heatmaps is an important future work.


\section{Conclusion}
\label{section: conclusion}

This paper proposed any-scale object detection using arbitrary-scale super-resolution(ASSR).
In addition to ASSR, an arbitrary-scale region proposal (ASRP) is proposed for suppressing false positive detections that occur due to ASSR.
The combination of ASSR and ASRP improves the memory efficiency compared with the previous method~\cite{DBLP:journals/access/AkitaU23} using discretized object scales.
Future work includes evaluating generalization to other datasets, exploring denser scale sampling, and testing different detectors and ASSR models.

This work was supported by JSPS under Grant Number Grant Number 22H03618.


\bibliographystyle{unsrt}
\bibliography{reference}

\end{document}